# Near-drowning Early Prediction Technique Using Novel Equations (NEPTUNE) for Swimming Pools

B David Prakash,
IAG Firemark Singapore,
davidprakash@gmail.com

**Abstract.** Safety is a critical aspect in all swimming pools. This paper describes a near-drowning early prediction technique using novel equations (NEPTUNE). NEPTUNE uses equations or rules that would be able to detect near-drowning using at least 1 but not more than 5 seconds of video sequence with no false positives. The backbone of NEPTUNE encompasses a mix of statistical image processing to merge images for a video sequence followed by K-means clustering to extract segments in the merged image and finally a revisit to statistical image processing to derive variables for every segment. These variables would be used by the equations to identify near-drowning. NEPTUNE has the potential to be integrated into a swimming pool camera system that would send an alarm to the lifeguards for early response so that the likelihood of recovery is high.

## 1. Introduction

The World Health Organization (WHO) classifies drowning as the 3$^{rd}$ leading cause of unintentional injury worldwide [1]. Globally, the highest drowning rates are among children aged between 1 to 4 years, followed by children aged between 5 to 9 years [1]. In individual countries such as the United States and France, within a short time frame of about 4 months, the number of drowning deaths in swimming pools and spas ranged between 74 to 163 [2-3]. Studies have shown that lifeguards may not be trained well enough to handle a drowning situation [4]. Hence, having a drowning detection system in conjunction with lifeguards in swimming pools would aid to promote swimming pool safety.

The existing drowning detection technologies can be broadly categorized into vision based systems [5-10] and wearable sensor based systems [11-13]. Vision based technologies can be further sub-categorized into those using underwater cameras [5-6] and those employing above water cameras [7-10]. A limitation of the use of underwater cameras is that they might miss the initial struggle that might take place above the water. Some drawbacks of the existing above water camera vision based technologies are that they have been demonstrated only using simulated video [7-10], they are trained to detect above water motionlessness [10] instead of the struggling motion which might pre-occur or might require additional costly fixtures such as a microarray to be mounted above the water to cover the entire swimming pool [7]. The shortcoming of a wearable based system is primarily the discomfort of use [8] which has an unproven possible notion that it might lead to younger children attempting to eliminate the discomfort by removing the device.

NEPTUNE is aimed at targeting the integration into existing above water camera(s) to enable a cost-effective installation by utilizing images from an existing camera fixture. It can identify pre-drowning struggling motions early using at least 1 but not more than 5 seconds of video sequence. The detection equations that NEPTUNE uses were derived from video sequences using an actual video footage [14-15] *.

*Please be informed that **Fig. 1.** contains confronting and real still images of a pre-drowning struggling victim.

## 2. NEPTUNE
### 2.1. Dataset

Two sets of videos were downloaded [14-15]. The first came with a manual red contour segmentation of the drowning victim throughout the entire video [14] and the second had the red contour segmentation of the drowning victim only in the initial portion of the video prior to the start of the pre-drowning struggle [15]. The second video [15] was used in the processing while the first [14] was applied as a confirmatory guide to locate the drowning victim in the second. Both videos were sequenced at 25 frames per second. Grey scale images were extracted from the second video [15] starting from the point of initial struggle and cropping was performed to output each image to a 447 by 281 dimension to maintain consistency with the camera coverage to the initial video. **Fig. 1.** shows a sample image extracted from [14] and another sample image extracted from [15] followed by grey scaling and cropping.

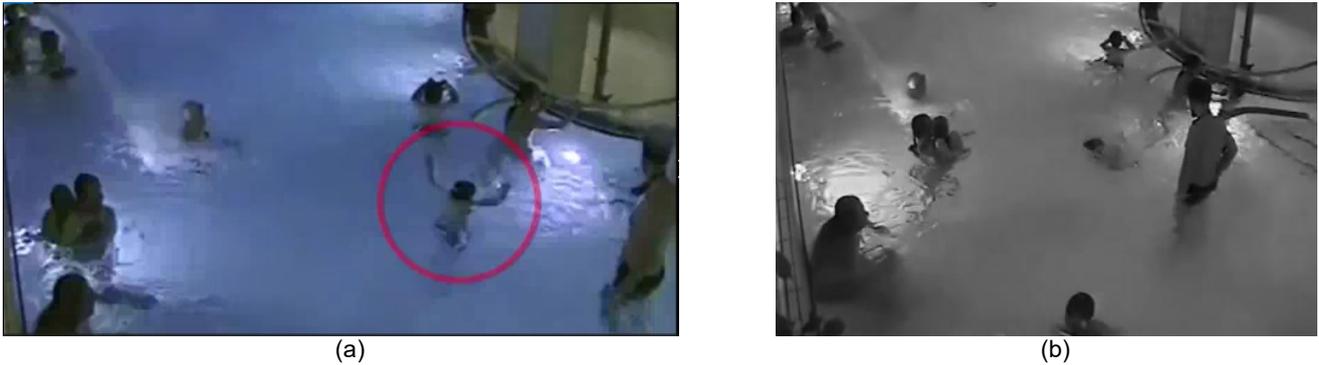

(a)        (b)

**Fig.1.** (a) Sample image extracted from [14] (b) Sample image extracted from [15] which had been grey scaled and cropped to follow similar camera coverage of the video from [14]

### 2.2. Pre-processing Pipeline

NEPTUNE's pre-processing pipeline consists of a combination of statistical image processing and K-means clustering [16].

The steps to process the images for every **m** seconds of video sequence are summarized below. The steps were repeated for five different values for **m** which are 1, 2, 3, 4 and 5.

1. Grey scaling (cropping was performed for the dataset used in this paper but would not be required in an actual setting)
2. Assuming 5 seconds of video sequences were being processed, for every pixel, the maximum absolute of the Fast Fourier Transform [17] across the 125 images was computed to give a two-dimensional matrix of 447 by 281. Since the video was sequenced at 25 frames per second, every 5 seconds would have a total 125 images that can be extracted. Similarly, every 4, 3, 2, and 1 seconds of video sequences would be using 100, 75, 50 and 25 images respectively.
3. The values across the two-dimensional matrix were normalized to a range between 0 and 1 to produce another two-dimensional matrix, **N**.
4. **N** would be transformed to a 1-dimensional array, **1-N** via repeated looping across the x-dimension shadowed by an inner loop across the y-dimension. For instance, the values at position (1,1), (1,2) … (1,447) of **N** would be placed at (1), (2) … (447) of **1-N** respectively while the values at position (2,1), (2,2) … (2,447) of **N** would be placed at (448), (449) … (894) respectively.
5. 3-means and 4-means clustering were performed independently on **1-N**. An earlier attempt of using 2-means and 3-means clustering resulted in the inability in finding segments either intersecting or close by the struggling victim for some positive video sequences. More segments were created using a 3-means and 4-means clustering and hence increased the probability of finding a segment that either intersects or is very close by the struggling victim for every positive video sequence. The purpose of having two types of clustering was to find pairs of nearest segments, one from $\{S_a\}$ and another from $\{S_b\}$ to have a larger pool of variables to explore.
    a. From the 3-means clustering, the largest of the 3 clusters was excluded using an assumption that it predominantly consisted of water. The 2 smaller clusters were remapped from **1-N** to **N** and a set of connected segments $\{S_a\}$ was extracted. A connected segment is a set of pixels belonging to the same cluster and each pixel within that segment had to be beside any one of the pixels in that segment.
    b. From the 4-means clustering, the largest of the 4 clusters was excluded using an assumption that it predominantly consisted of water. The 3 smaller clusters were remapped from **1-N** to **N** and a second set of connected segments $\{S_b\}$ was extracted.

**Fig. 2.** shows an example of how 3-means clusters would look after remapping **1-N** back to **N**. In comparison to **Fig. 1.**, the cluster with the majority of pixels belonged to water for **Fig. 2a. Fig. 2b.** illustrates an example of a connected segment.

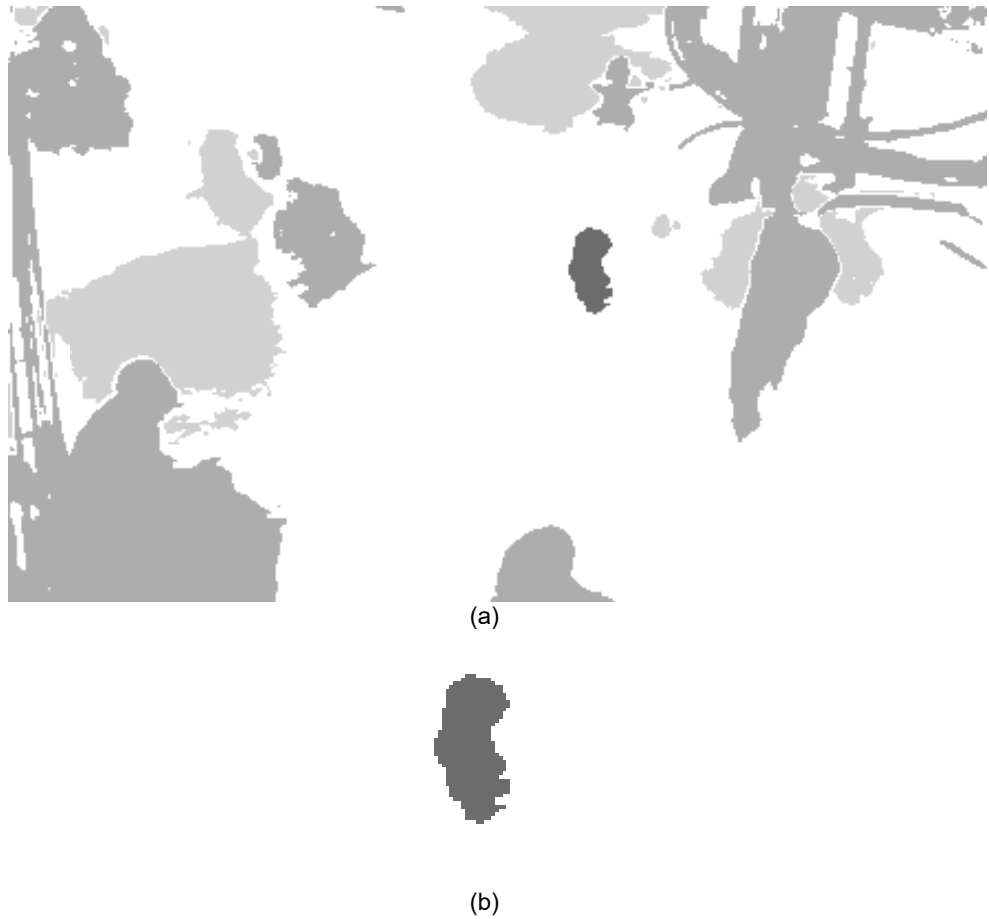

(a)

(b)

**Fig.2.** (a) 3 clusters created from 3-mean clustering. Water is predominantly within the largest cluster which is shaded in white. The two shades of grey represent the remaining two clusters. The connected segment coloured in black intersects with the struggling victim and belongs to the cluster with the darker shade of grey. (b) A zoomed in image of the single connected segment coloured in black.

6. For each segment in **{$S_a$}**, variables shown in **Table 1** were derived. Variable **V1** was reserved for the labelling of presence/absence of a pre-drowning struggling victim within/close by a segment. If there was no segment that intersects the struggling victim, the nearest segment would be considered to contain the pre-drowning struggling victim.

**Table 1.** Variables derived for each segment in **{$S_a$}**

| Variable Name | Variable Description |
|---|---|
| **V2** | Ratio of V4 to V3 |
| **V3** | Number of pixels in the segment |
| **V4** | Standard deviation of values from 5 points (4 extreme points* and the segment's centre) |
| **V5** | Ratio of V2 to the sum of V2 across all segments |
| **V6** | Ratio of V3 to the sum of V3 across all segments |
| **V7** | Ratio of V4 to the sum of V4 across all segments |

*4 extreme points are {minimum(x), minimum(y)}, {minimum(x), max(y)}, {maximum(x), minimum(y)} and {maximum(x), maximum(y)} where x and y are the x-coordinates and y-coordinates of a segment respectively

7. For each segment in **{S<sub>a</sub>}**, similar variables for the respective nearest segment from **{S<sub>b</sub>}** was computed as shown in **Table 2**.

Table 2. Variables derived from respective nearest segment in {S<sub>b</sub>}

| Variable Name | Variable Description |
|---|---|
| V8 | Ratio of V10 to V9 |
| V9 | Number of pixels in the segment |
| V10 | Standard deviation of values of 5 points (4 extreme points* and the segment's centre) |
| V11 | Ratio of V11 to the sum of V11 across all segments |
| V12 | Ratio of V12 to the sum of V12 across all segments |
| V13 | Ratio of V13 to the sum of V13 across all segments |

*4 extreme points are {minimum(x), minimum(y)}, {minimum(x), maximum(y)}, {maximum(x), minimum(y)} and {maximum(x), maximum(y)} where x and y are the x-coordinates and y-coordinates of a segment respectively

8. Next, for each segment in **{S<sub>a</sub>}**, new variables were created as shown in **Table 3**. These new variables were computed using the variables derived from **Table 1** and **Table 2**.

Table 3. Variables derived from respective nearest segment in {S<sub>b</sub>}

| Variable Name | Variable Description |
|---|---|
| V2_8 | Ratio of V2 to V8 |
| V3_9 | Ratio of V3 to V9 |
| V4_10 | Ratio of V4 to V10 |
| V5_11 | Ratio of V5 to V11 |
| V6_12 | Ratio of V6 to V12 |
| V7_13 | Ratio of V7 to V13 |

9. Finally, the percentile cut-offs for every variable across all segments in **{S<sub>a</sub>}** were computed as shown in **Tables A1 (a), (b), (c)** and **(d)** of the **Appendix**.
Every variable value for each segment would be assigned either a value of 1, 2, 3 or 4 according to the range which they fall into as indicated in **Table 4** with respect to length of video sequence. For example, say for a segment in a 5s video sequence having a **V2** value of 0.03, it would be transformed to a value of 2 since it is between the 25$^{th}$ and 50$^{th}$ percentiles of **V2**.

Table 4. Value assigned each variable in every segment in {S<sub>a</sub>}

| Value Assigned | Range |
|---|---|
| 1 | being less than or equal to the 25$^{th}$ percentile |
| 2 | being more than the 25$^{th}$ percentile but less than or equal to the 50$^{th}$ percentile |
| 3 | being more than the 50$^{th}$ percentile but less than or equal to the 75$^{th}$ percentile |
| 4 | being more than the 75$^{th}$ percentile |

Solely for this training, labelling of **V1** for every segment had to be performed. It would contain one of two values; either 1 if it was positive or 0 otherwise. A positive refers to the presence of a struggling pre-drowning victim within or close by the segment. This labelling would not be required for the actual application of NEPTUNE. The video sequence lengths of 1s, 2s, 3s, 4s and 5s respectively contained 36, 17, 12, 9, and 8 positive video sequences. A positive video sequence would each entail a struggling pre-drowning victim. Correspondingly, for the video sequence lengths of 1s, 2s, 3s, 4s and 5s, the respective number of video sequences for which there were no struggling were 946, 406, 269, 179 and 138. Each video sequence may have multiple segments with at most 1 positive segment.

## 2.3. Equations Derivation

The equations were derived from optimized rules generated via association rules mining [18]. Association rules mining was attempted as an approach to detect the positives as anomalies. The existence of a non-linear relationship between the variables and the presence/absence of positives is shown in **Fig. 3.** where there is a poor correlation not exceeding 0.37 between the actual and predicted scores for all the video sequence lengths studied**.** This further justifies the use of association rules mining since it can identify both linear and non-linear properties that would distinguish positives from non-positives. The formulae of the regression model used in **Fig. 3a, b, c, d** and **e** are shown in **Table A2** of the **Appendix**.

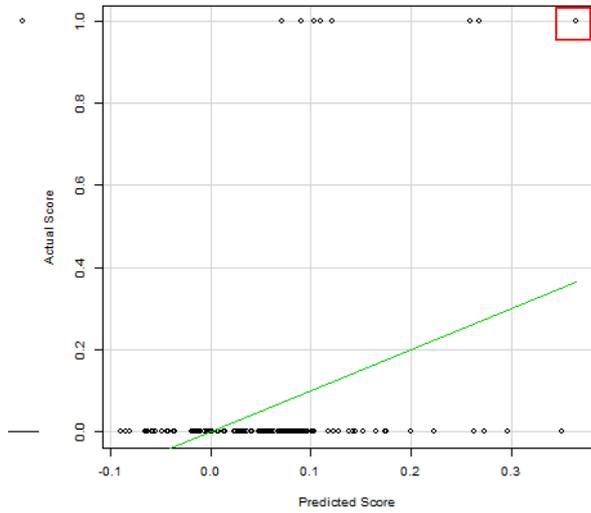

(a) Video Sequence Length of 5s

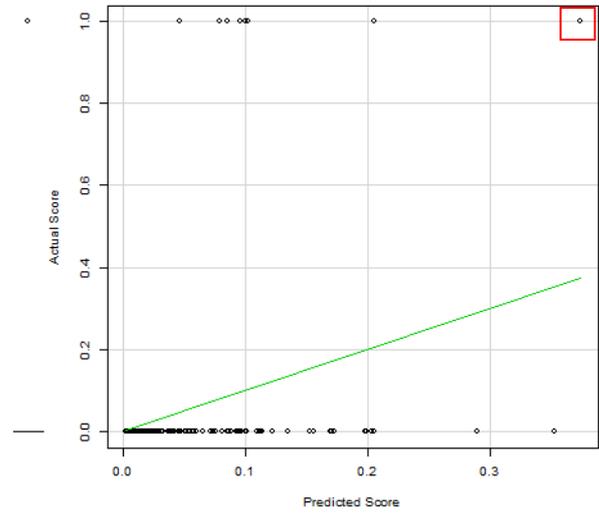

(b) Video Sequence Length of 4s

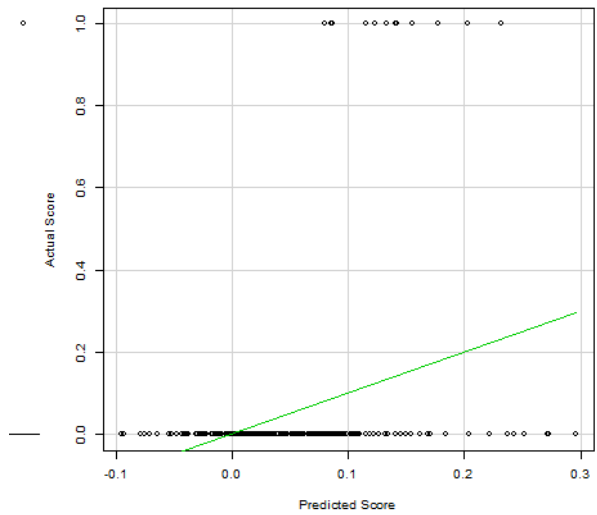

(c) Video Sequence Length of 3s

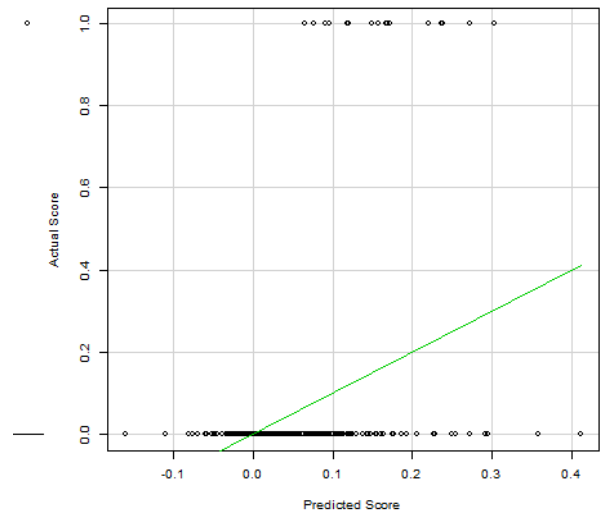

(d) Video Sequence Length of 2s

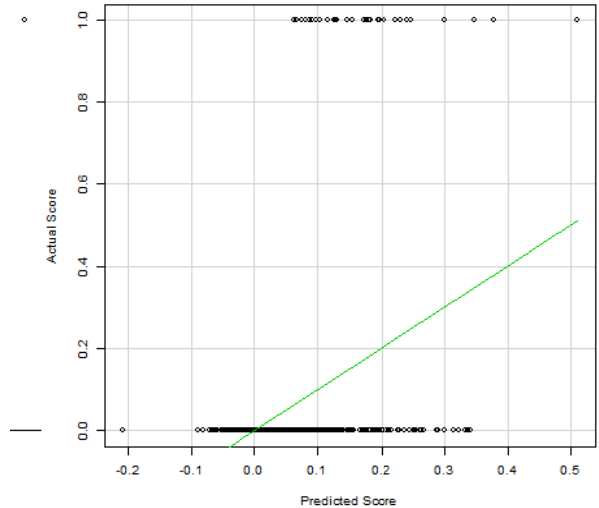

(e) Video Sequence Length of 1s

**Fig. 3.** The predicted scores created via a AIC stepwise linear regression model [19] built using the entire dataset against the actual scores for the various video sequence lengths. A score of 1 denotes a positive. Positive segments with the highest predicted scores for video sequence lengths of 4s and 5s are each enclosed within a red bounding box.

For video sequence lengths of 4s and 5s, the positive segments with the highest predicted score had the largest and most circular segment as shown respectively in **Fig. 4.** and **Fig. 5.** which made these two positive segments more distinguishable than the other positive segments within the respective video sequence lengths.

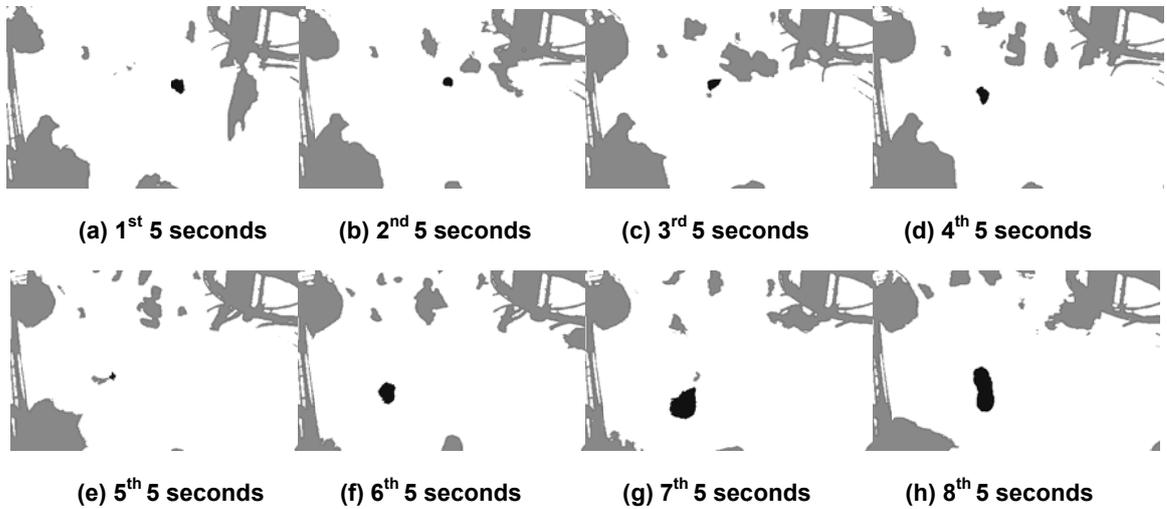

(a) 1st 5 seconds  (b) 2nd 5 seconds  (c) 3rd 5 seconds  (d) 4th 5 seconds

(e) 5th 5 seconds  (f) 6th 5 seconds  (g) 7th 5 seconds  (h) 8th 5 seconds

**Fig. 4.** The positive segments shaded in black for the first 8 video sequences of length 5s where there was a struggling victim. The 7th 5 seconds video sequence had the highest prediction score using linear regression.

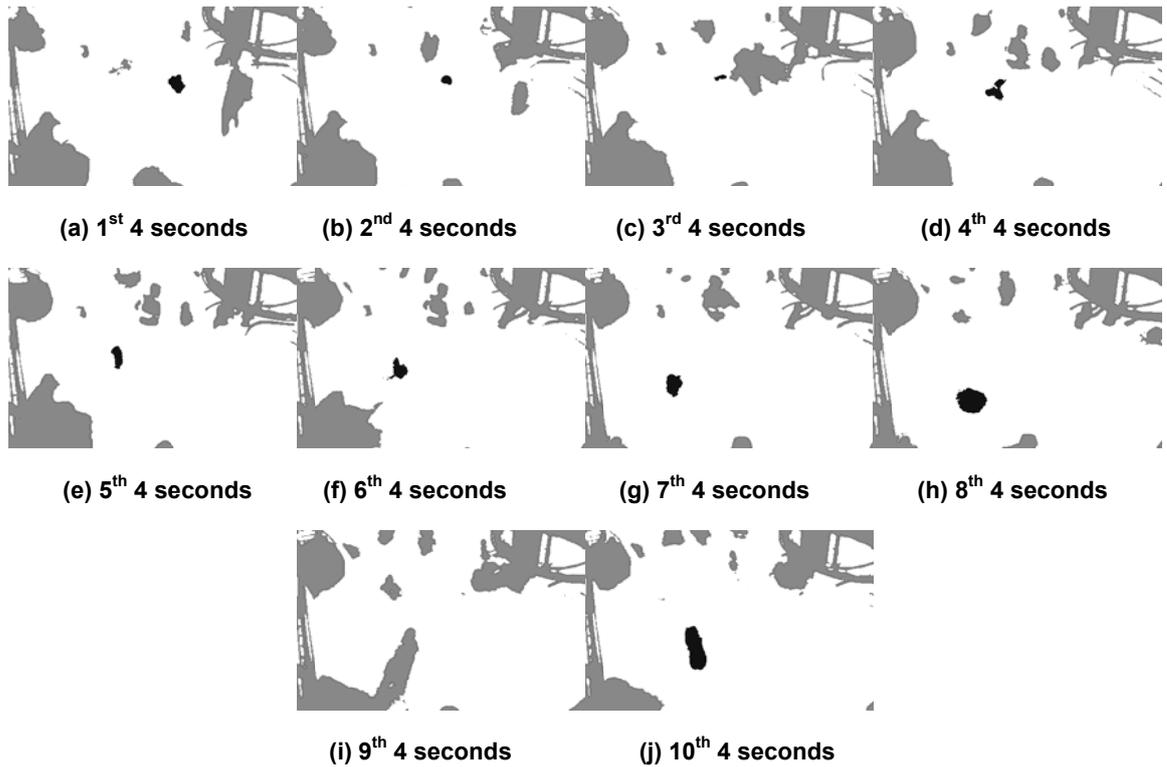

**Fig. 5.** The positive segments shaded in black for the first 10 video sequences of length 4s where there was a struggling victim. The 8th 4 seconds video sequence had the highest prediction score using linear regression. No positive segment was detectable in 9th 4 seconds video sequence although there was a struggling.

Rules were generated independently for the various video sequence lengths to detect presence/absence of a positive segment using 19 variables of which 18 came from **Tables 1, 2** and **3**. One of the 19 variables was used for the labelling of a absence/presence of a positive. All rules were generated for a confidence of 1 and targeted a minimum of 1 positive. We generated different rule sets across different number of variables ranging from 3 to the full 19. For every number of variables used for rule set generation, one had to be the label. For instance, if 3 variables were used, 2 would be derived variables and 1 would be the label. **Fig. 6.** showsthat the number of number of positives identifiable for various number of variables used across the various video sequence lengths.

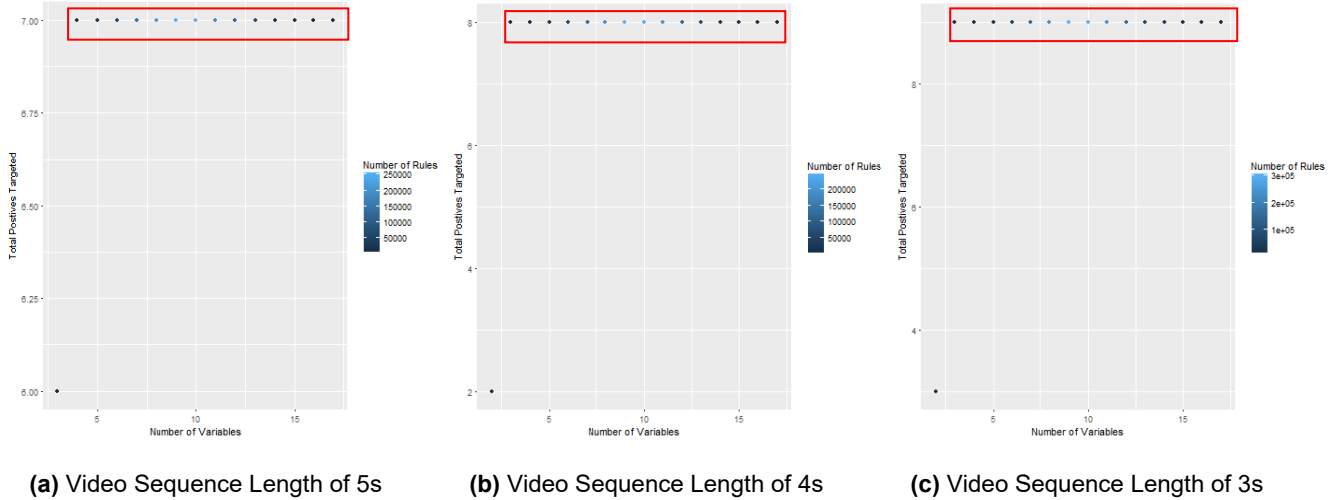

(a) Video Sequence Length of 5s  (b) Video Sequence Length of 4s  (c) Video Sequence Length of 3s

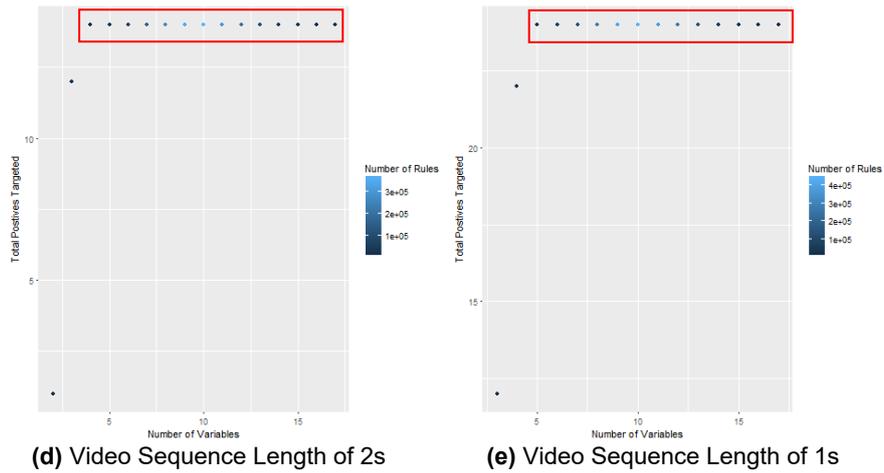

**(d)** Video Sequence Length of 2s    **(e)** Video Sequence Length of 1s

**Fig. 6.** Total positives targeted and number of rules generated with 100% confidence in detecting positives across the corresponding number of variables used in the rules generation. Each spot represents a rule set consisting of rules generated for the respective number of variables. Rule sets which targeted the maximum number of positives detectable for the respective video sequences lengths were enclosed within a red bounding box

The proportion of the rules generated using variables that detected the maximum number of positives either targeting 2 or 3 positives across the different video sequence lengths are shown in **Fig. 7**. Having a higher proportion of individual rules targeting more positives would ensure that the rule set is more generic and hence applicable across other datasets. Therefore, for the final rule set in each of the video sequence lengths studied, rules generated with the highest proportion of individual rules targeting the most number of positives would be chosen. It can be seen in **Fig. 7.** that the rules sets with the highest proportion of rules targeting 2 or 3 positives tend to be generated when less variables were used. Using less variables would reduce the specificity of the rules generated thereby allowing more generic rules to be produced. Each rule in final rule set would be henceforth referred to as an equation.

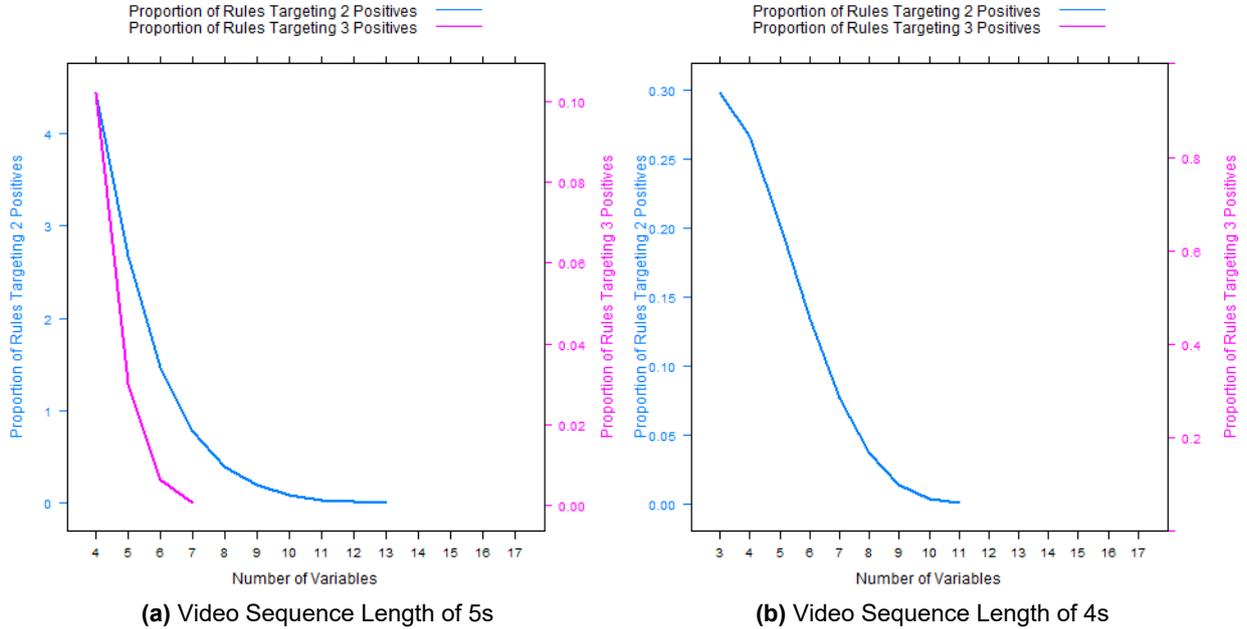

**(a)** Video Sequence Length of 5s    **(b)** Video Sequence Length of 4s

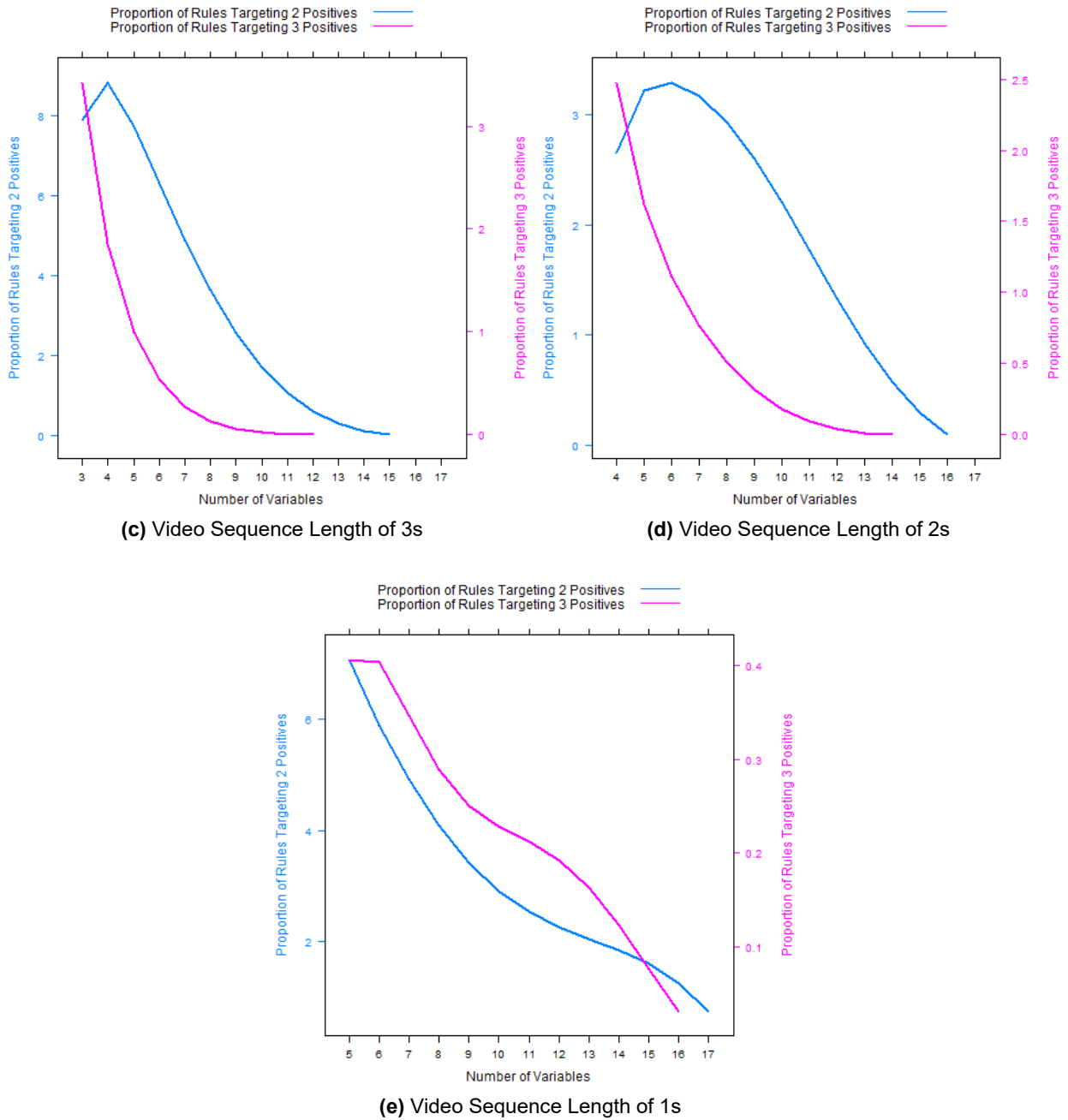

**(c)** Video Sequence Length of 3s

**(d)** Video Sequence Length of 2s

**(e)** Video Sequence Length of 1s

**Fig. 7.** Proportion of rules targeting 2 or 3 positives across the various number of variables used for rules generation

## 3. Results

**Fig. 8.** shows the positive detections during the 40s of struggle. It should be noted there were time points when only one of the video sequence lengths detected the presence of a struggle. For instance, at the 39[th] second of struggle, only the video sequence length of 3s could detect the struggle. To elaborate further, the detection of a struggle at the 39[th] second of struggle for a video sequence of length 3s relied on the video sequence from the 36[th] second of struggle up to the 39[th] second of struggle. Hence, it is proposed to use all the video sequence lengths in parallel for pre-drowning struggling detection to maximize the detectable time points of struggle.

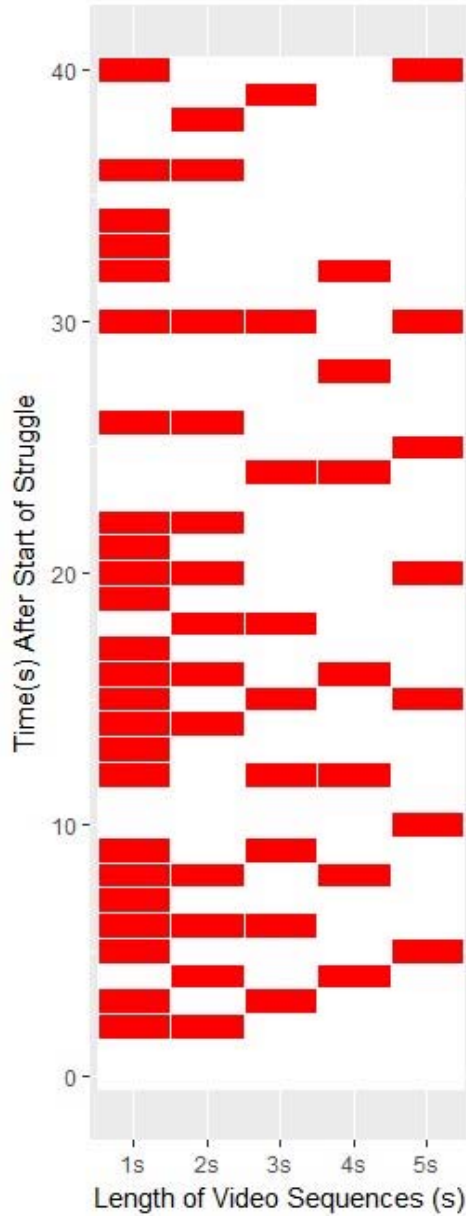

**Fig. 8.** Pre-drowning detection throughout the 40s of struggle for the various video sequence lengths. Positive detection only occurs at the time points where there is a red rectangle. For instance, at 40s after the start of struggle, only video sequence lengths of 1s and 5s were able to detect the presence of a struggle.

4. Conclusion

NEPTUNE was presented in this paper as a feasible technique for early detection of struggling pre-drowning victims. The reliance on solely camera based video sequences would allow an easy integration into existing infrastructure in swimming pools with camera(s). This is the first vision based technique built using real video sequences and the fastest requiring at least 1 but not more than 5 seconds of video footage for detection. With collaborative initiatives to collate more actual pre-drowning video footages, NEPTUNE can be refined and further developed into a real-time cost-effective vision based early pre-drowning detection system.

Acknowledgments

This work has been supported by IAG Firemarks Singapore. I like to thank everyone in the IAG Firemarks Singapore team for their continued encouragement especially Allison Howells for her initial and continued cheering of what started as a simple possibility to a prospective reality.

Appendix

**Table A1 (a).** Percentile cut-offs for each variable across all segments in {S$_a$} using 5s of video sequences

| Variable Name | 25th Percentile | 50th Percentile | 75th Percentile |
| --- | --- | --- | --- |
| V2 | 0.025537 | 0.089672 | 0.136088 |
| V3 | 4 | 14.5 | 232.5 |
| V4 | 0.616663 | 1.83083 | 5.959598 |
| V5 | 0.033195 | 0.079342 | 0.160908 |
| V6 | 0.002247 | 0.012159 | 0.144687 |
| V7 | 0.017574 | 0.054455 | 0.176347 |
| V8 | 0.03157 | 0.095445 | 0.197203 |
| V9 | 5 | 23 | 187.75 |
| V10 | 0.817409 | 2.087051 | 6.102764 |
| V11 | 0.001802 | 0.005235 | 0.012307 |
| V12 | 0.000165 | 0.000753 | 0.006355 |
| V13 | 0.002531 | 0.005998 | 0.015984 |
| V2_8 | 0.543143 | 0.808972 | 1.39471 |
| V3_9 | 0.38125 | 1.294801 | 2.176598 |
| V4_10 | 0.475384 | 1.027277 | 1.501146 |
| V5_11 | 7.80108 | 14.34464 | 29.0312 |
| V6_12 | 5.501654 | 15.77497 | 50.7646 |
| V7_13 | 4.08387 | 9.045111 | 18.00813 |

**Table A1 (b).** Percentile cut-offs for each variable across all segments in {S$_a$} using 4s of video sequences

| Variable Name | 25th Percentile | 50th Percentile | 75th Percentile |
| --- | --- | --- | --- |
| V2 | 0.028521 | 0.096291 | 0.181219 |
| V3 | 4 | 13 | 175.5 |
| V4 | 0.552183 | 1.605571 | 5.200332 |
| V5 | 0.036251 | 0.094939 | 0.158282 |
| V6 | 0.002502 | 0.011834 | 0.130142 |
| V7 | 0.017525 | 0.055982 | 0.15279 |
| V8 | 0.028882 | 0.092357 | 0.197203 |
| V9 | 4 | 14 | 258.5 |
| V10 | 0.555836 | 1.951793 | 6.336658 |
| V11 | 0.001904 | 0.006055 | 0.013018 |
| V12 | 0.000128 | 0.000523 | 0.008301 |
| V13 | 0.001747 | 0.00508 | 0.017826 |
| V2_8 | 0.645632 | 0.89624 | 1.350195 |
| V3_9 | 0.5 | 1.12129 | 2.333333 |
| V4_10 | 0.560464 | 1.005624 | 1.50856 |
| V5_11 | 6.285207 | 12.07338 | 27.94023 |
| V6_12 | 7.657875 | 15.97658 | 56.00498 |
| V7_13 | 5.056071 | 9.226469 | 21.33621 |

**Table A1 (c).** Percentile cut-offs for each variable across all segments in {S$_a$} using 3s of video sequences

| Variable Name | 25th Percentile | 50th Percentile | 75th Percentile |
|---|---|---|---|
| V2 | 0.03194 | 0.096123 | 0.150733 |
| V3 | 4 | 13 | 159 |
| V4 | 0.555836 | 1.566484 | 5.153229 |
| V5 | 0.030702 | 0.082569 | 0.143622 |
| V6 | 0.002227 | 0.01207 | 0.106299 |
| V7 | 0.018207 | 0.047486 | 0.136789 |
| V8 | 0.027022 | 0.096123 | 0.197203 |
| V9 | 3 | 11 | 212 |
| V10 | 0.394405 | 1.457435 | 6.193818 |
| V11 | 0.001662 | 0.005614 | 0.011111 |
| V12 | 0.000108 | 0.000351 | 0.007508 |
| V13 | 0.001214 | 0.003857 | 0.015949 |
| V2_8 | 0.588914 | 0.856317 | 1.418486 |
| V3_9 | 0.5 | 1.402542 | 2.5 |
| V4_10 | 0.706932 | 1.084431 | 1.927521 |
| V5_11 | 6.548546 | 13.00316 | 29.72699 |
| V6_12 | 8.128327 | 19 | 64.53311 |
| V7_13 | 5.06461 | 10.99971 | 23.4315 |

**Table A1 (d).** Percentile cut-offs for each variable across all segments in {S$_a$} using 2s of video sequences

| Variable Name | 25th Percentile | 50th Percentile | 75th Percentile |
|---|---|---|---|
| V2 | 0.027442 | 0.096123 | 0.181219 |
| V3 | 4 | 11 | 185 |
| V4 | 0.544352 | 1.31045 | 5.561496 |
| V5 | 0.032004 | 0.081686 | 0.140064 |
| V6 | 0.001807 | 0.006944 | 0.110295 |
| V7 | 0.015953 | 0.039461 | 0.137062 |
| V8 | 0.028312 | 0.096123 | 0.197203 |
| V9 | 4 | 15 | 232 |
| V10 | 0.555836 | 1.972027 | 6.934729 |
| V11 | 0.001601 | 0.004623 | 0.010108 |
| V12 | 0.000116 | 0.000536 | 0.007775 |
| V13 | 0.00136 | 0.004605 | 0.016561 |
| V2_8 | 0.653128 | 0.897169 | 1.275932 |
| V3_9 | 0.5 | 1.25969 | 2.084053 |
| V4_10 | 0.502151 | 1.019212 | 1.580454 |
| V5_11 | 8.305886 | 13.30195 | 28.31558 |
| V6_12 | 6.32996 | 16.15734 | 48.24208 |
| V7_13 | 4.833126 | 10.26422 | 23.45575 |

**Table A1 (d).** Percentile cut-offs for each variable across all segments in {S$_a$} using 1s of video sequences

| Variable Name | 25th Percentile | 50th Percentile | 75th Percentile |
|---|---|---|---|
| V2 | 0.043662 | 0.102924 | 0.197203 |
| V3 | 3 | 8 | 109.75 |
| V4 | 0.394405 | 1.099948 | 4.615003 |
| V5 | 0.032882 | 0.074918 | 0.12968 |
| V6 | 0.001575 | 0.006651 | 0.073862 |
| V7 | 0.01319 | 0.034193 | 0.119747 |
| V8 | 0.039327 | 0.111167 | 0.197203 |
| V9 | 3 | 10 | 123 |
| V10 | 0.547526 | 1.344413 | 5.361706 |
| V11 | 0.001796 | 0.005085 | 0.008561 |
| V12 | 0.000106 | 0.000323 | 0.004121 |
| V13 | 0.001244 | 0.003122 | 0.012171 |
| V2_8 | 0.043662 | 0.102924 | 0.197203 |
| V3_9 | 3 | 8 | 109.75 |
| V4_10 | 0.394405 | 1.099948 | 4.615003 |
| V5_11 | 0.032882 | 0.074918 | 0.12968 |
| V6_12 | 0.001575 | 0.006651 | 0.073862 |
| V7_13 | 0.01319 | 0.034193 | 0.119747 |

**Table A2.** Regression formulae for the respective models in Fig 3.

| Fig | Formula |
|---|---|
| 3a | $V1 = -0.6882 * V2 + 0.8153 * V6 - 1.1143 * V7 - 0.0003874 * V9 + 13.0915 * V12 - 0.001769 * V6\_12 + 0.002818 * V7\_13 + 0.1227$ |
| 3b | $V1 = 0.01103 * V4 + 0.00107 * V15 - 0.00123$ |
| 3c | $V1 = -0.3232 * V2 + 0.0001431 * V3 - 0.1553 * V6 - 0.2502 * V8 - 0.09636$ |
| 3d | $V1 = 0.01665 * V4 - 0.1753 * V6 - 0.2300 * V8 + 0.0002441 * V9 - 7.6024 * V12 - 0.005745 * V4\_10 + 0.03580$ |
| 3e | $V1 = -0.2072 * V2 + 0.01528 * V4 - 0.1563 * V6 + 0.000243 * V9 - 0.01994 * V10 - 2.9549 * V11 - 7.4156 * V12 + 7.6710 * V13 - 0.004826 * V16 + 0.58134$ |